\documentclass[10pt,twocolumn,letterpaper]{article} 

\usepackage{cvpr}              

\usepackage[font=small,labelsep=period]{caption}
\usepackage{subcaption}
\usepackage{graphicx}
\usepackage{amsmath}
\usepackage{amssymb}
\usepackage{multirow}
\usepackage{booktabs}
\usepackage{array}
\usepackage[normalem]{ulem}
\usepackage{tabularx}
\usepackage[accsupp]{axessibility}  

\newcolumntype{Y}{>{\centering\arraybackslash}X}
\newcolumntype{R}{>{\raggedleft\arraybackslash}X}
\newcolumntype{L}{>{\raggedright\arraybackslash}X}

\usepackage[pagebackref=true, breaklinks, colorlinks, linkcolor=red, citecolor=blue, urlcolor=blue]{hyperref}

\usepackage[capitalize]{cleveref}
\crefname{section}{Sec.}{Secs.}
\Crefname{section}{Section}{Sections}
\Crefname{table}{Table}{Tables}
\crefname{table}{Tab.}{Tabs.}

\newcommand{\campose}{\mathcal{T}}
\newcommand{\ray}{\textbf{r}}
\newcommand{\loss}{\mathcal{L}}

\newcolumntype{P}[1]{>{\centering\arraybackslash}p{#1}}

\newcommand\customparagraph[1]{\vspace{0.6em}\noindent\textbf{#1}}

\usepackage{xcolor}
\usepackage{etoolbox}

\newif\ifshowedits

\newcommand{\addeditor}[3]{%
  \definecolor{#1color}{rgb}{#3}
  \expandafter\newcommand\csname #1\endcsname[1]{%
  \ifshowedits
    {\color{#1color} ##1}%
  \else
    {##1}%
  \fi
  }%
  \expandafter\newcommand\csname #1rmk\endcsname[1]{%
  \ifshowedits
    {\color{#1color} {\bf [#2: ##1]}}
  \fi
  }%
  \expandafter\newcommand\csname #1rpl\endcsname[2]{%
  \ifshowedits
    {\color{#1color} ##1 \sout{##2}}
  \else
    {##1}
  \fi
  }%
}

\newcommand{\mycomment}[1]{}

\newcommand{\createtextvar}[1]{
  \expandafter\newcommand\csname #1\endcsname{%
  {\text{#1}}
}%
}

\newcommand{\textvars}[1]{\forcsvlist{\createtextvar}{#1}}

\newcommand{\calH}{{\cal H}}

\newcommand{\calM}{{\cal M}}
\newcommand{\calN}{{\cal N}}

\newcommand{\calR}{{\cal R}}
\newcommand{\calS}{{\cal S}}
\newcommand{\calT}{{\cal T}}

\newcommand{\bd}{{\bf d}}

\newcommand{\bh}{{\bf h}}

\newcommand{\bn}{{\bf n}}

\newcommand{\br}{{\bf r}}

\newcommand{\bx}{{\bf x}}

\DeclareMathOperator*{\argmin}{arg\,min}

\renewcommand{\ie}{\emph{i.e.}}
\renewcommand{\eg}{\emph{e.g.}}

\addeditor{vincent}{VL}{0.0, 0.5, 0.0}
\addeditor{shreyas}{SH}{0.0, 0.0, 0.0}
\addeditor{tomas}{TH}{0.0, 0.0, 1.0}
\addeditor{luan}{LT}{0.1, 0.1, .7}
\showeditstrue

\def\cvprPaperID{702} 
\def\confName{CVPR}
\def\confYear{2023}

\begin{document}




\title{In-Hand 3D Object Scanning from an RGB Sequence}
\newcommand{\namesep}{\hspace{0.8em}}
\author{Shreyas Hampali$^{1,3}$\thanks{Work done as part of Shreyas's PhD thesis at TU Graz, Austria.}~\namesep
Tomas Hodan$^{1}$\namesep
Luan Tran$^{1}$\\
Lingni Ma$^{1}$\namesep
Cem Keskin$^{1}$\namesep
Vincent Lepetit$^{2,3}$ \and
\textsuperscript{1}{\normalsize Reality Labs at Meta}\\ 
\textsuperscript{2}{\normalsize 
LIGM, Ecole des Ponts, Univ Gustave Eiffel, CNRS, Marne-la-Vall\'ee, France
}\\
\textsuperscript{3}{\normalsize Institute for Computer Graphics and Vision, Graz University of Technology, Graz, Austria}\\
{\tt\small Project Website: \href{https://rgbinhandscanning.github.io/}{https://rgbinhandscanning.github.io/}}
}

\maketitle

\begin{abstract}
We propose a method for in-hand 3D scanning of an unknown object with a monocular camera. Our method relies on a neural implicit surface representation that captures both the geometry and the appearance of the object, however, by contrast with most NeRF-based methods, we do not assume that the camera-object relative poses are known. Instead, we simultaneously optimize both the object shape and the pose trajectory. As direct optimization over all shape and pose parameters is prone to fail without coarse-level initialization, we propose an incremental approach that starts by splitting the sequence into carefully selected overlapping segments within which the optimization is likely to succeed. We  reconstruct the object shape and track its poses independently within each segment, then merge all the segments before performing a global optimization. We show that our method is able to reconstruct the shape and color of both textured and challenging texture-less objects, outperforms classical methods that rely only on appearance features, and that its performance is close to recent methods that assume known camera poses.
\vspace{-2.5ex}

\end{abstract}


\section{Introduction}

\begin{figure}[h!]
    \begin{center}
    \includegraphics[trim=80 165 600 75, clip, width=0.94\columnwidth]{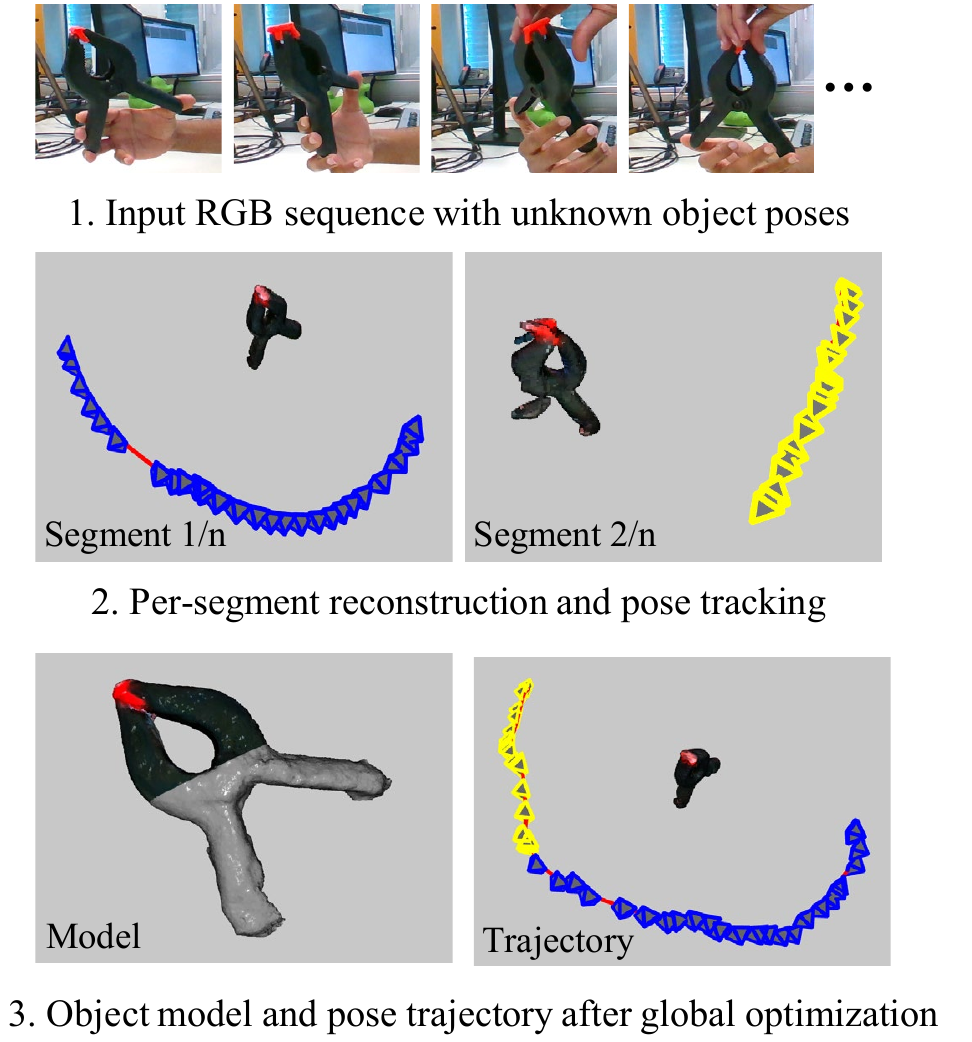}
    \vspace{-0.5ex}
    \captionof{figure}{Given an RGB sequence of a hand manipulating an unknown object, our method reconstructs the 3D shape and color of the object, even if the object surface is non-Lambertian or poorly textured. We first split the input sequence into multiple overlapping segments (two in this figure) in which the object can be reliably reconstructed and tracked. We then use the tracked object-camera relative poses to initialize a global optimization that produces the final object model and refined pose trajectory.
    } \label{fig:teaser}
    \end{center}
    \vspace{-3.0ex}
\end{figure}

Reconstructing 3D models of unknown objects from multi-view images is a
computer vision problem which has received considerable attention~\cite{han2019image}. With a single camera, a user can capture multiple views of an object
by manually moving the camera around a static object~\cite{runz2020frodo, unisurf, yariv2020multiview} or by turning the object in front of the camera~\cite{rusinkiewicz-02-realtime3dmodelacquisition, Weise, Weise2, tzionas-iccv15-3dobjectreconstruction}. 
\shreyas{The latter approach is often referred to as \textit{in-hand object scanning} and is convenient for reconstructing objects from cameras mounted on an AR/VR headset such as Microsoft HoloLens or Meta Quest. Moreover, this approach can reconstruct the full object surface, including the bottom part which cannot be scanned in the static-object setup.}


Recent 3D reconstruction methods rely on neural representations~\cite{park-cvpr19-deepsdf, mescheder2019occupancy, yariv2020multiview, unisurf, DVR, yariv2021volume}. By contrast with earlier reconstruction methods~\cite{hartley-00-multipleviewsgeometry}, the recent methods can provide an accurate dense 3D reconstruction even in non-Lambertian conditions and without any prior knowledge of the object shape. 
However, most of these methods assume that the camera poses are provided, typically by Structure-from-Motion~(SfM) methods such as COLMAP~\cite{schonberger2016pixelwise}. 
Applying SfM methods to in-hand object scanning is problematic as these methods require a sufficient number of distinct visual features and can thus handle well only textured objects. NeRF-based methods such as~\cite{barf, scnerf, nerfmm, neroic}, which simultaneously estimate the radiance field of the object and the camera poses without requiring initialization from COLMAP, are restricted to forward-facing camera captures. As we experimentally demonstrate, these methods fail to converge
if the images cover a larger range of viewpoints, which is typical for in-hand scanning.

We propose a method for in-hand object scanning from an RGB image sequence with unknown camera-object relative poses.
We rely on a neural representation that captures both the geometry and the appearance of the object and therefore enables reconstructing even poorly textured objects, as shown in Fig.~\ref{fig:teaser}.
By contrast with most NeRF-based methods, we do not assume that the camera poses are available and instead simultaneously optimize both the object model and the camera trajectory. As global optimization over all input frames is prone to fail, we propose an incremental optimization approach. We start by splitting the sequence into carefully selected overlapping segments within which the optimization is likely to succeed. We then optimize our objective for incremental object reconstruction and pose tracking within each segment independently. The segments are then combined by aligning poses estimated at the overlapping frames, and we finally optimize the objective globally over all frames of the input sequence to achieve complete object reconstruction.

Note that we do not make any assumptions on the type of hand-object grasps and also consider scenarios where the grasping is dynamic, i.e.,  contact points continuously change, which corresponds to natural hand-object interactions. This is in contrast with the recent work \cite{huang2022hhor} that considers only static grasps. In fact, we refrain from considering hand poses in our method as they cannot be reliably estimated under occlusion in case of dynamic grasps, which could lead to incorrect object reconstruction.

We experimentally demonstrate that the proposed method is able to reconstruct the shape and color of both textured and challenging texture-less objects.
We evaluate the method on datasets HO-3D~\cite{hampali-cvpr20-honnotate}, RGBD-Obj~\cite{wang_dataset} and on the newly captured sequences with challenging texture-less objects.
We show that the proposed method
achieves higher-quality reconstruction than COLMAP~\cite{schonberger2016pixelwise}, which fails to estimate the object poses in the case of poorly textured objects and is in par with a strong baseline method which uses ground-truth object poses. Our method also outperforms a very recent single-image based object reconstruction method~\cite{cmu}, even though this method is trained on sequences of the same object. We also demonstrate the real-world applicability of our method by in-hand scanning an object with ARIA glasses~\cite{aria_pilot_dataset} (see supplement).
\section{Related Work}


This section reviews previous methods for in-hand scanning and general object reconstruction from color images, and compares them with the method proposed in this paper.

\subsection{In-Hand Object Scanning}

Using an RGB-D sensor, several in-hand scanning systems~\cite{rusinkiewicz-02-realtime3dmodelacquisition,Weise,Weise2,Weise3,Wang2021DemoGraspFL} rely on tracking and are able to recover the shape of small objects manipulated by hands. Later, \cite{tzionas-iccv15-3dobjectreconstruction} showed how to use the motion of the hand and its contact points with the object to add constraints useful to deal with texture-less and highly symmetric objects, while restricting the contact points to stay fixed during the scanning. Unfortunately, the requirement for an RGB-D sensor limits applications of these techniques.

More recently, with the development of deep learning, several methods have shown that it is possible to infer the object shape from a single image~\cite{hasson-cvpr19-learningjointreconstruction,karunratanakul-20-graspingfield,cmu} after training on images of hands manipulating an object with annotations of the object pose and shape. Given the fact that the geometry is estimated from a single image, the results are impressive. However, the reconstruction quality is still limited, especially because these methods do not see the back of the object and cannot provide a good prediction of the appearance of the object for all possible viewpoints. In this paper, we propose an approach for in-hand object scanning which estimates the shape and color of a completely unknown object from a sequence of RGB images, without any pre-training on annotated images.

\subsection{Reconstruction from Color Images}

Recovering the 3D geometry of a static scene and the camera poses from multiple RGB images has a long history in computer vision~\cite{faugeras-93-book,hartley-00-multipleviewsgeometry,snavely2006photo,furukawa2009accurate}. Structure-from-Motion (SfM) methods are now very robust and accurate, however they are limited to scenes with textures, which is not the case for many common objects. 

In the past few years, with the emergence of neural implicit representations as effective means of modeling 3D geometry and appearance, many methods~\cite{DVR, unisurf, nerf, yariv2020multiview, yariv2021volume} reconstruct a 3D scene by optimizing a neural implicit representation from multi-view images by minimising the discrepancy between the observed and rendered images. These methods achieve impressive reconstructions on many scenes, but they still need near perfect camera poses, which are typically estimated by Structure-from-Motion methods.

Several NeRF-based methods have attempted to retrieve the camera poses while reconstructing the scene.  Methods such as NeRF-\,\!-~\cite{nerfmm}, SCNeRF~\cite{scnerf} and BARF~\cite{barf} show that camera poses can be estimated even when initialized with identity matrix or random poses while simultaneously estimating the radiance field. However, these methods are shown to converge only on forward facing scenes and require coarse initialization of poses for $360^{\circ}$ captures as in in-hand object scanning. More recently, SAMURAI~\cite{samurai}) used manual rough quadrant annotations for coarse-level pose initialization and showed that object shape and material can be recovered along with the accurate camera poses.


In this work we propose to estimate the camera-object relative pose in from RGB image sequence and reconstruct the object shape without any prior information of the object or its poses. Unlike previous methods, we rely on the temporal information and incrementally reconstruct the object shape and estimate its pose.

\textvars{bg,col,seg,BCE,opf,reg,inc,trac,OF,dep}

\renewcommand{\col}{\text{c}}
\renewcommand{\seg}{\text{s}}
\renewcommand{\opf}{\text{f}}
\renewcommand{\reg}{\text{r}}
\renewcommand{\dep}{\text{d}}

\section{Proposed Method} \label{sec:method}

In this section, we first describe the considered setup and how we represent the object with a neural representation. Then, we derive an objective function for estimating the object reconstruction and the camera pose trajectory, and explain how we optimize this function.


\subsection{In-Hand Object Scanning Setup}
\label{sec:reconstruction_setup}

\paragraph{Input and Output.} Our input is a sequence of RGB images showing an unknown rigid object being manipulated by one or two hands in the field of view of the camera. The output is a color 3D model of the manipulated object.
The input sequence is captured by an egocentric camera or a camera mounted on a tripod. In both cases, the relative pose between the camera and the object is unknown. In order to achieve full reconstruction of the object, the image sequence is assumed to show the object from all sides.

\shreyas{\customparagraph{Available Information About Objects and Hands.} The segmentation masks of the object and hands are assumed available for all input images. In our experiments, we obtain the masks by of-the-shelf networks -- by Detic~\cite{zhou2022detecting}, which can segment unknown objects in a single RGB image, or by DistinctNet~\cite{boerdijk2020learning} which can segment an unknown moving object from a pair of images with static background.
We additionally use segmentation masks from SeqFormer~\cite{wu2021seqformer} to ignore pixels of hands that manipulate the object.
}

\customparagraph{Phong Reflection Model and Distant Lights.} The object to reconstruct is assumed to be solid (\ie, non-translucent), and we model the reflectance properties of the object surface with the Phong reflection model~\cite{HughesDamEtAl13} and assume that the light sources are far from the object and the camera. Under the Phong model, the observed color at a surface point depends on the viewing direction, the surface normal direction, and the light direction. If the light sources are far, the incoming light direction can be approximated to remain unchanged, which allows us to use the standard neural radiance field~\cite{nerf} to model the object appearance. This assumption is reasonable as rotation is the primary transformation of the object during in-hand manipulation--on the HO-3D dataset, which contains sequences of a hand manipulating objects, the maximum standard deviation of the object's 3D location is only 7.9cm.





\subsection{Object Representation} 
\label{sec:object_representation}

\noindent\textbf{Implicit Neural Fields.} As in UNISURF~\cite{unisurf}, we represent the object geometry by an occupancy field and the object appearance by a color field, with each realized by a neural network. The occupancy field
is defined as a mapping: $o_\theta(\bx): \mathbb{R}^3 \rightarrow [0,1]$,
%
%
where $\theta$ represents the parameters of the network and $\bx$ is a 3D point in the object coordinate system. The object surface is represented by 3D points $\calS = \{\bx\,|\, o_\theta(\bx)=0.5\}$,
and the surface mesh can be recovered by the Marching Cubes algorithm~\cite{mcubes}.

The color field is a mapping: $c_\theta(\bx; \bd, \bn, \bh): \mathbb{R}^3 \times \mathbb{R}^3 \times \mathbb{R}^3 \times \mathbb{R}^n \rightarrow \mathbb{R}^3$ that represents the color at a surface point $\bx \in \calS$ and is conditioned on the viewing direction $\bd$ (\ie, the direction from the camera center to the point $\bx$), the normal vector $\bn$ at $\bx$, and the geometry feature $\bh$ at $\bx$ which has $n$ dimensions and is extracted from the occupancy field network. The color for a particular pixel/ray $\br$ is defined as $\hat{C}_i(\ray) = c_\theta(\bx_s)$, where $\bx_s$ is the closest point on the object surface along ray $\br$ (the object is assumed non-translucent). 
To simplify the notation, we include in $\theta$ both the parameters of the occupancy field and of the color field as these two networks are optimized together.

\customparagraph{Rendering.}
As in~\cite{unisurf}, the rendered color at a pixel in a frame $i$ is obtained by integrating colors along the ray $\ray$ originating from the camera center and passing through the pixel. The continuous integration is approximated as:
\begin{gather}
    \hat{C}_i(\ray) = \sum_{k=1}^M \gamma(\bx_k) c_\theta(\bx_k; \bd_k, \bh_k, \bn_k)  \\
    \text{with  } \gamma(\bx_k) = o_\theta(\bx_k)\prod_{l<k}\big(1-o_\theta(\bx_l)\big) \> ,
    \label{eq:rend_unisurf}
\end{gather}
where $\{\bx_k\}$ are $M$ samples along the ray $\ray$. The alpha-blending coefficient $\gamma(\bx_k)$, is defined as in~\cite{unisurf}, is $1$ if point $\bx_k$ is on the visible surface of the object and $0$ otherwise.

\subsection{Reconstruction Objective}

In UNISURF~\cite{unisurf}, the network parameters $\theta$ are estimated by solving the following optimization problem:
%
\begin{gather}
    \theta^* = \argmin_\theta \sum_i \sum_{\ray \in \calR_i} \loss_{\col}^i(\ray) \> ,\\
    \loss_{\col}^i(\ray) = \lvert\lvert \hat{C}_i(\ray) - C_i(\ray)\lvert\lvert \>,
    \label{eq:loss_col}
\end{gather}
where $\loss_{\col}^i(\ray)$ is the photometric loss measuring the difference between the rendered color $\hat{C}_i(\ray)$ and the observed color $C_i(\ray)$ at a pixel intersected by the ray $\ray$ in the frame $i$, and $\calR_i$ is the set of rays sampled in the frame $i$.


In our case, we additionally
optimize the camera poses:
%
\begin{align}
    \label{eq:complete_loss}
    \theta^*, \{\campose_i^*\} = \argmin_{\theta, \{\campose_i\}} \sum_i \Big( \sum_{\ray\in \calH_i} \loss_{\col}^i(\ray)+\!\!\!\sum_{\ray\in \calM_i} \loss_{\seg}^i(\ray)\Big) \> ,
\end{align}
where $\campose_i$ is the camera pose of frame $i$ expressed by a rigid transformation from the camera coordinate system to the object coordinate system. $\calH_i$ is the set of object rays in frame $i$, and $\calM_i$ is the set of object and background rays in frame $i$.
We only use rays passing through the object and background pixels and ignore the hand pixels. The term $\loss_{\seg}^i(\ray)$
is a segmentation loss for ray $\ray\in\calM_i$:
\begin{align}
    \label{eq:seg_loss}
    \loss_{\seg}^i(\ray) &= \BCE\Big(\max_k~\{o_\theta(\bx_k)\}, S_i(\ray) \Big) \> ,
\end{align}
where $\BCE(\cdot)$ is the binary cross-entropy loss, and $S_i(\ray)$ is the object mask value for ray $\ray$ in the frame $i$ (the mask is obtained as described in Sec.~\ref{sec:reconstruction_setup}). The value of $S_i(\ray)$ is 1 if the pixel corresponding to ray $\ray$ lies in the provided object mask, and 0 otherwise. The term $\max_k~\{o_\theta(\bx_k)\}$ is the maximum occupancy along the ray $\ray$ according to the estimated occupancy field $o_\theta(.)$, and is expected to be 1 if $\ray$ intersects the object and 0 otherwise.


\subsection{Optimization}
\label{sec:opt_obj}

Directly optimizing Eq.~\eqref{eq:complete_loss} is prone to fail. As we show in Sec.~\ref{sec:experiments}, a random (or a fixed) initialization of poses followed by an optimization procedure similar to the one used in BARF~\cite{barf} leads to degenerate solutions.
Instead, we propose an incremental optimization approach which starts by splitting the sequence into carefully selected overlapping segments, within which the optimization is more likely to succeed (Sec.~\ref{sec:splitting}). We optimize the objective in each segment by incremental frame-by-frame reconstruction and tracking, with the objective being extended by additional loss terms to stabilize the tracking (Sec.~\ref{sec:segment_tracking}). Then, we merge the segments by aligning poses estimated at the overlapping frames (Sec.~\ref{sec:stitch}), and finally optimize the objective globally over all frames of the sequence (Sec.~\ref{sec:stitch}).

\begin{figure}[t]
    \centering
    \includegraphics[trim=0 185 0 0, clip,width=0.98\linewidth]{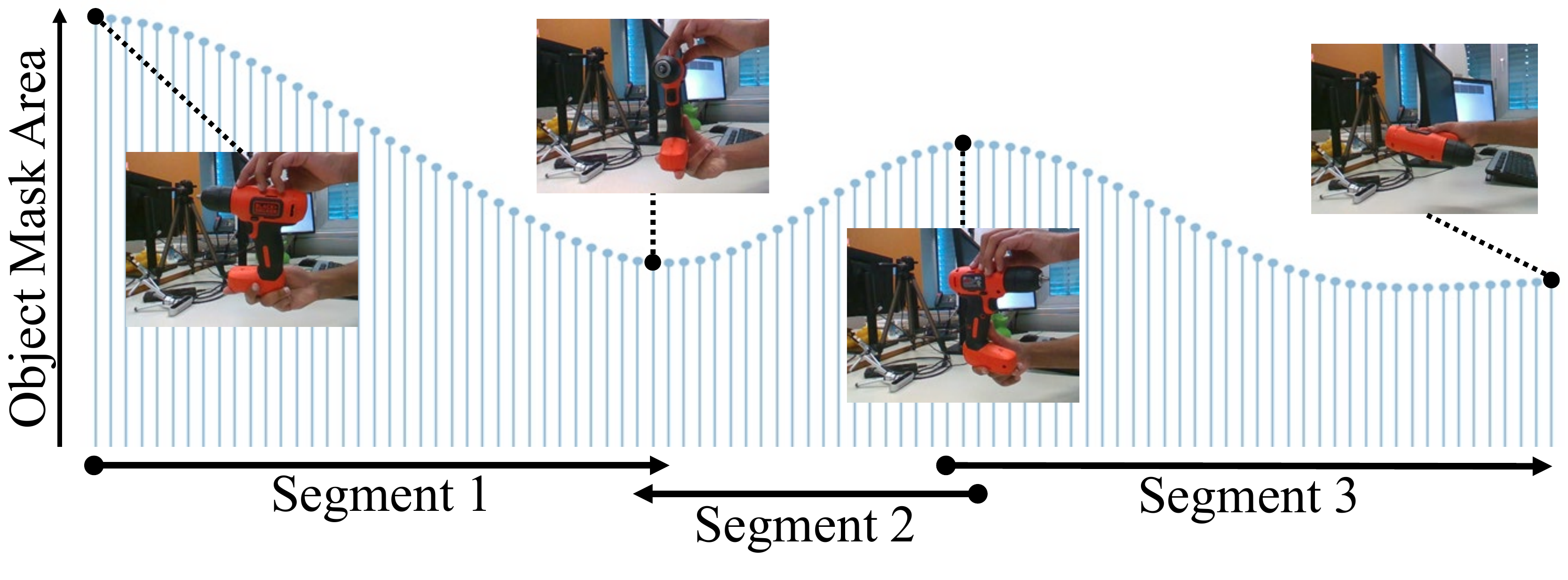}
    \vspace{-1ex}
    \caption{\textbf{Splitting a sequence into easy-to-track segments.} The segment boundaries are defined at frames with locally maximal or minimal area of the object mask (the start and the end of each segment is shifted by a few frames from the extremum to make the segments overlap). Note that we can track backwards in time, from a local maximum to a local minimum.}
    \label{fig:seg_area}
    \vspace{-2ex}
\end{figure}

\subsubsection{Input Sequence Segmentation}
\label{sec:splitting}

We observed in our early experiments that frame-to-frame tracking is prone to fail when 
previously observed parts of the object start disappearing and new parts start appearing. This is not surprising as there is no 3D knowledge about the new parts yet, and the current reconstruction of the object is disappearing and cannot be used to track these new parts.

We therefore propose to split the sequence into segments so that tracking on each segment is unlikely to drift much. How can we detect when new parts are appearing? We observe that this can be done based on the apparent area\footnote{As mentioned earlier, we obtain a mask of the object by segmenting the image, so we can easily compute its apparent area.} of the object: Under the assumption that the distance of the object to the camera and the occlusions by the hand do not change much, large parts of the object disappear when the apparent area goes through a minimum (see Figure~\ref{fig:seg_area}).
We therefore split the input sequence into multiple segments, with their boundaries defined at frames where the apparent area reaches local maxima and minima, and process each segment from the local maximum to the local minimum (\eg, segments 1 and 3 in Figure~\ref{fig:seg_area} are processed from left to right and segment 2 is processed from right to left). The local extrema are computed from the smoothed curve of the apparent object area using a sliding window with the length of 12 frames.
The start and the end of each segment is shifted by a few frames from the extremum to introduce overlaps with the neighboring segments (the overlaps are used in Sec.~\ref{sec:stitch} to merge the estimated per-segment pose trajectories).
With this approach, tracking within a segment starts with a point of view where the object reprojection area is large in the image, which facilitates bootstrapping of the tracking together with our shape regularization loss.

\subsubsection{Per-Segment Optimization}
\label{sec:segment_tracking}


Within each segment, we iteratively optimize the following objective on a progressively larger portion of the segment allowing us to incrementally reconstruct the object and track its pose. The $T$ frames in a segment are denoted by the set $\{\calS\}_{i=1}^T$. Over the course of the optimization, the index $t$ of the currently considered frame progresses from the first to the last frame of the segment, and for each step we solve:
%
%
\begin{align}
\label{eq:inc_loss}
    &\theta^*, \{\campose_i^*\}_{i=1}^t = \argmin_{\theta, \{\campose_i\}_{i=1}^t} \quad \sum_{i=1}^t  \sum_{\ray\in\calH_i} \loss_\col^i (\ray) + \text{...} \\ & \sum_{i=1}^t \sum_{\ray\in\calM_i} \Big(\loss_\seg^i(\ray) + \loss_\opf^i(\ray) + \loss_\reg^i(\ray)\Big) \;+ \sum_{i=1}^{t-1}\sum_{\ray\in\calM_i}\loss_\dep^i(\ray) \> . \nonumber
\end{align}
%
The terms $\loss_\col^i$ and $\loss_\seg^i$ are the color and mask losses defined in Eq.~\eqref{eq:complete_loss}, $\loss_\opf^i$ is a loss based on optical flow that provides constraints on the poses, $\loss_\reg^i$ is a shape regularization term that prevents degenerate object shapes, and $\loss_\dep^i$ is a synthetic-depth loss that stabilizes the tracking. 
$\calH_i$ is the set of rays going through pixels on the object in frame $i$, and 
 $\calM_i$ the set of rays going through pixels on the object or the background in frame $i$.
More details on the loss terms $\loss_\opf^i, \loss_\reg^i$ and $\loss_\dep^i$ are provided later in this section.

The network parameters $\theta$ are initialized by their estimate from the previous iteration $t-1$. The camera pose $\calT_t$ for $t>1$ is initialized using the second-order motion model applied to the previous poses $\{\calT_i\}_{i=0}^{t-1}$. The first camera pose $\calT_0$ is initialized to a fixed distance from the origin of the object coordinate system and orientated such that the image plane faces the origin. At each iteration, we sample a fixed percentage of rays from the new frame (set to $15\%$ empirically) and the rest from the previous frames.

\begin{figure}[h!]
    \begin{center}
    \includegraphics[width=0.8\linewidth]{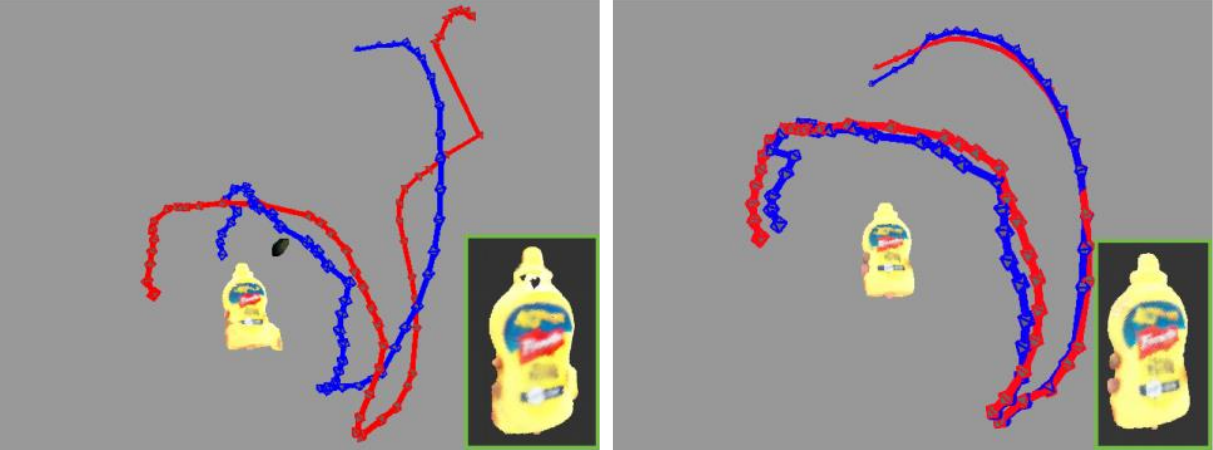}\vspace{-1ex}
    \caption{\textbf{Effect of the optical flow loss $\loss_\opf$.} Pose estimates (red) are more stable when the loss $\loss_\opf$ is applied (right). The ground-truth poses are shown in blue.}
    \label{fig:optical_flow}
    \vspace{-3ex}
    \end{center}
\end{figure}

\customparagraph{Optical Flow Loss.} This term provides additional constraints on the camera poses and is defined as:

\noindent \resizebox{0.94\linewidth}{!}{
\begin{minipage}{\linewidth}
\vspace{-0.5mm}
\begin{eqnarray}
\loss_\opf^i(\ray) = \sum_k \gamma(\bx_k) \big( \pi_i(\bx_k) - \pi_{i\text{-}1}(\bx_k)  - \OF_i(\pi_{i\text{-}1}(\bx_k)) \big)^2 \> ,
\end{eqnarray}
\vspace{0.5mm}
\end{minipage}
}

\noindent where $\{\bx_k\}_i$ are 3D points along ray $\ray$, $\pi_i(\bx)$ is the 2D reprojection of the point $\bx$ in the frame $i$, $\OF_i$ is the optical flow between frame $i-1$ and frame $i$, and $\gamma(\cdot)$ is as defined in Eq.~\eqref{eq:rend_unisurf} and evaluates to one for points on the object surface and zero elsewhere. Fig.~\ref{fig:optical_flow} shows the effect of optical flow loss on the trajectory after several optimization steps.
We use
\cite{yang2019vcn} to compute the optical flow.



\begin{figure}[h!]
    \begin{center}
    \includegraphics[width=\linewidth]{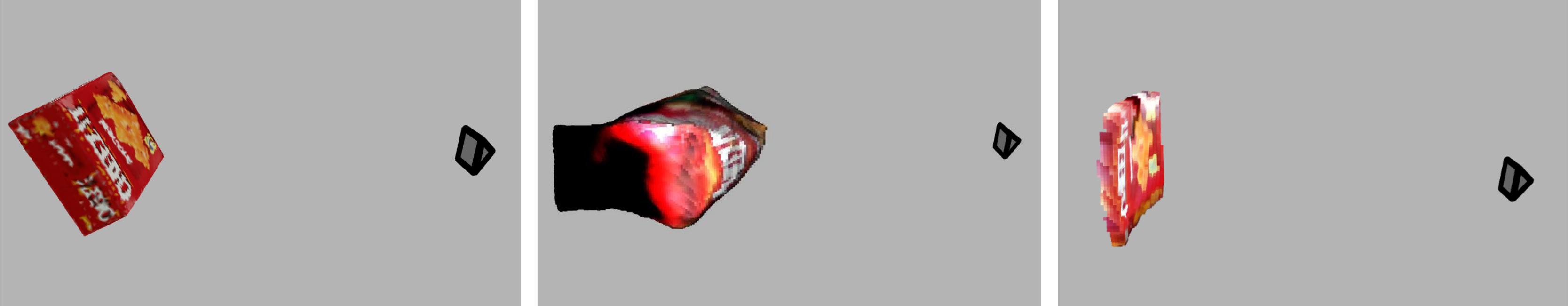}\vspace{-1ex}
    \caption{\textbf{Effect of the shape regularization loss $\loss_\reg$.} Left to right: The ground-truth object mesh, implicit surface reconstructed without the regularization term at $t=1$, and implicit surface reconstructed with the regularization term at $t=1$.
    }
    \label{fig:regularization}
    \vspace{-3ex}
    \end{center}
\end{figure}

\customparagraph{Shape Regularization Loss.} During early iterations~(\ie, when $t$ is small), the occupancy field is under-constrained and needs to be regularized to avoid degenerate object shapes. We introduce a regularization that encourages reconstruction near the origin of the object coordinate system:
\begin{equation}
\label{eq:reg_loss}
    \loss_\reg^i(\ray) = \sum_k o_\theta(\bx_k) \exp{(\alpha \cdot \|\bx_k\|_2)} \> ,
\end{equation}
where $\alpha$ is a hyperparameter. At $t=0$, minimizing $\loss_\reg^i$ results in an object surface that is parallel to the image plane (see supplement for explanation).
Encouraging a planar proxy as an approximation of the object shape helps to stabilize the early stage of the optimization. Fig.~\ref{fig:regularization} shows examples achieved with and without the regularization.

\begin{figure}[h!]
    \begin{center}
        \includegraphics[width=0.8\linewidth]{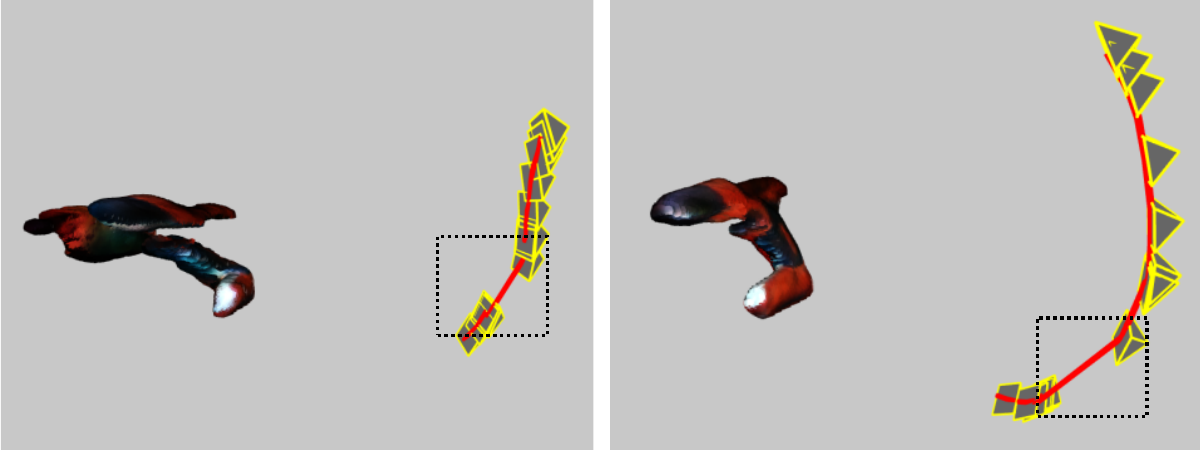}\vspace{-1ex}
    \caption{\textbf{Effect of the synthetic-depth loss $\loss_\dep$.} Large pose changes (highlighted by black boxes) can deform previously reconstructed parts of the object if the depth loss $\loss_\dep$ is not used (left).
    }
    \label{fig:depth_term}
    \end{center}
    \vspace{-4ex}
\end{figure}

\customparagraph{Synthetic-Depth Loss.} We also introduce a loss based on synthetic depth maps rendered by the object shape estimate. The motivation for this term is to regularize the evolution of the shape estimate and prevent its drift. It is defined as:

%
%
\begin{equation}
    \loss_\dep^i (\ray) = \big(\sum_k \gamma(\bx_k) \text{dep}_i(\bx_k)-\hat{d}_i(\ray)\big)^2 \> ,
\end{equation}
where $\text{dep}_i(\bx_k)$ is the depth of the point $\bx_k$ along the ray $\ray$, $\gamma(\cdot)$ is as defined in Eq.~\ref{eq:rend_unisurf} and 
$\hat{d}_i$ is the depth map rendered using the previous estimates of the object model and the camera pose for frame $i$.
Note that $\loss_\dep^i$ is only applied on rays from frames $[1,t-1]$ at optimization step $t$ of Eq.~\eqref{eq:inc_loss} whose synthetic depths are pre-computed.  Figure~\ref{fig:depth_term} illustrates the contribution of this term.

\subsubsection{Global Optimization}
\label{sec:stitch}

The camera trajectories and object reconstruction for each segment are recovered up to a rigid motion and a scaling factor. To express the overlapping segments in a common coordinate frame, we align the pose estimates at the overlapping frames with the following procedure.
Let $\campose_i^k =[\phi_i, t_i]$ be the rotation and translation of the camera for the frame $i$ in the segment $k$ with $N_s$ frames. We obtain a normalized pose by taking $\hat{\campose}_i^k = \big[\phi_i; t_i/\frac{1}{N_s}\sum_j \|t_j\|\big]$. 
We then retrieve the rigid motion $\campose_{k_1\rightarrow k_2}$~(rotation and translation) that aligns two overlapping segments $k_1$ and $k_2$:
\begin{equation}
    \campose_{k_1\rightarrow k_2} = \argmin_\campose \sum_i \|\campose\cdot\hat{\campose}_i^{k_1}-\hat{\campose}_{\calN(i)}^{k_2}\|_F \> ,
\end{equation}
where $\|\cdot\|_F$ denotes the Frobenius norm, $\calN(i)$ is the frame index in segment $k_2$ corresponding to frame $i$ in segment $k_1$, and the summation is over the set of all overlapping frames. In practice, we observed that as less as a single overlapping frame is sufficient for connecting the segments.

\begin{figure}[]
    \begin{center}
    \begin{minipage}{.33\linewidth}
    \subcaptionbox{From random poses}
    {\includegraphics[trim=180 230 640 175, clip,width=1\linewidth]{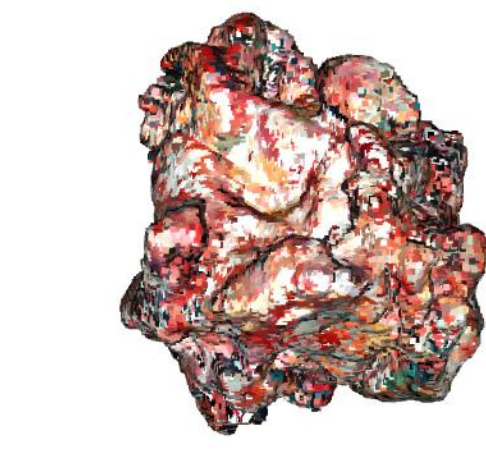}}
    \end{minipage}~
    \begin{minipage}{.33\linewidth}
    \subcaptionbox{From the same pose}
    {\includegraphics[trim=320 155 350 100, clip,width=1\linewidth]{figures/ablation/init_ablation.pdf}}
    \end{minipage}~
    \begin{minipage}{.33\linewidth}
    \subcaptionbox{From our pose est.}
    {\includegraphics[trim=600 210 180 150, clip,width=1\linewidth]{figures/ablation/init_ablation.pdf}}
    \end{minipage}
    \vspace{0.5ex}
    \caption{\textbf{Reconstruction from different initial poses.} Only initialization from coarse pose estimates yield a meaningful solution.}
    \label{fig:rand_init}
    \end{center}
    \vspace{-2ex}
\end{figure}

\begin{figure}[]
    \begin{center}
    \includegraphics[trim=70 300 230 80, clip,width=0.93\linewidth]{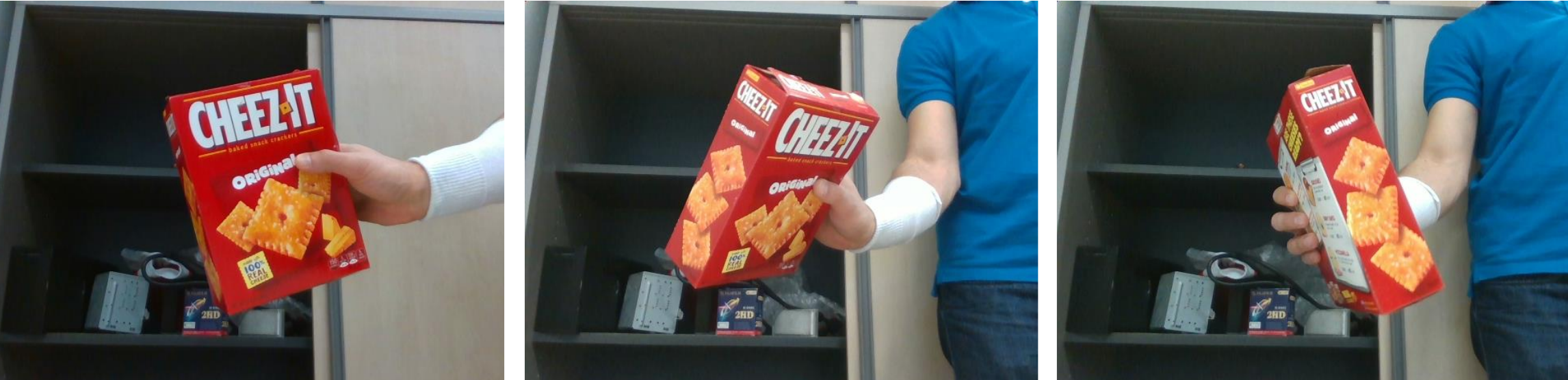}\\
    \includegraphics[trim=70 130 230 250, clip,width=0.93\linewidth]{figures/ablation/segments/segment_ablation.pdf}\vspace{-1ex}
    \caption{\textbf{Incremental reconstruction and pose tracking is prone to fail beyond the segment boundary.}
    On this representative example, the incremental reconstruction and pose tracking procedure works well as long as the front face is visible. When the front face starts to disappear and new parts start to appear (last column), reconstruction degrades and pose tracking drifts.}
    \label{fig:seg_ablation}
    \end{center}
    \vspace{-2.5ex}
\end{figure}

We use the aligned poses of two neighboring segments as pose initialization and optimize the objective function from Eq.~\eqref{eq:complete_loss}. The network parameters $\theta$ are initialized to reconstruct a sphere. The neighboring segments are combined iteratively until we obtain complete reconstruction from the full sequence. Fig.~\ref{fig:rand_init} shows the reconstruction with different pose initializations -- even for a textured object, coarse initialization is necessary for convergence.
In Fig.~\ref{fig:seg_ablation}, we show a situation where the incremental reconstruction and pose tracking continues beyond the segment boundary -- the solution degrades when new surface parts appear.

\vspace{-1.5ex}

\section{Implementation Details}
The occupancy and color field networks are implemented by 8-layer MLP's with ReLU activations and a hidden dimension of $F$. Fourier features~\cite{nerf} at $k_x$ octaves are used to encode the 3D coordinates, and at $k_d$ octaves to encode the view direction. During the per-segment optimization, similar to~\cite{lasr}, instead of directly optimizing the 6D pose parameters, the pose is parameterized with a CNN that takes the RGB image as input and outputs the 6DoF pose. Weights of the CNN are initialized with weights pre-trained on ImageNet~\cite{imagenet}. The CNN provides a neural basis for the pose parameters and acts as a regularizer. Without the CNN parameterization, the per-segment optimization procedure described in the section Sec.~\ref{sec:segment_tracking} typically fails. 

During the per-segment optimization (Sec.~\ref{sec:segment_tracking}), we set $F$\,$=$\,$128$, $k_x$\,$=$\,$4$, $k_d$\,$=$\,$2$ and run 6k gradient descent iterations at each tracking step. Further, at each step, we add 5 frames to increase the optimization speed. 

For the global optimization (Sec.~\ref{sec:stitch}), we use $F$\,$=$\,$256$, $k_x$\,$=$\,$8$, $k_d$\,$=$\,$4$ and run 25k gradient descent iterations for a pair of segments. We use smooth masking of frequency bands as described in BARF~\cite{barf} for better convergence and optimize the 6D pose variables directly instead of using CNN parameterization in this stage.
The frames are subsampled such that their maximum number
is $150$.

We compute the local maxima and minima from the object area curve as explained in Sec.~\ref{sec:splitting} by first  performing a Gaussian filtering of per frame object areas.

\begin{figure*}[t!]
\begin{center}
\includegraphics[width=0.95\linewidth]{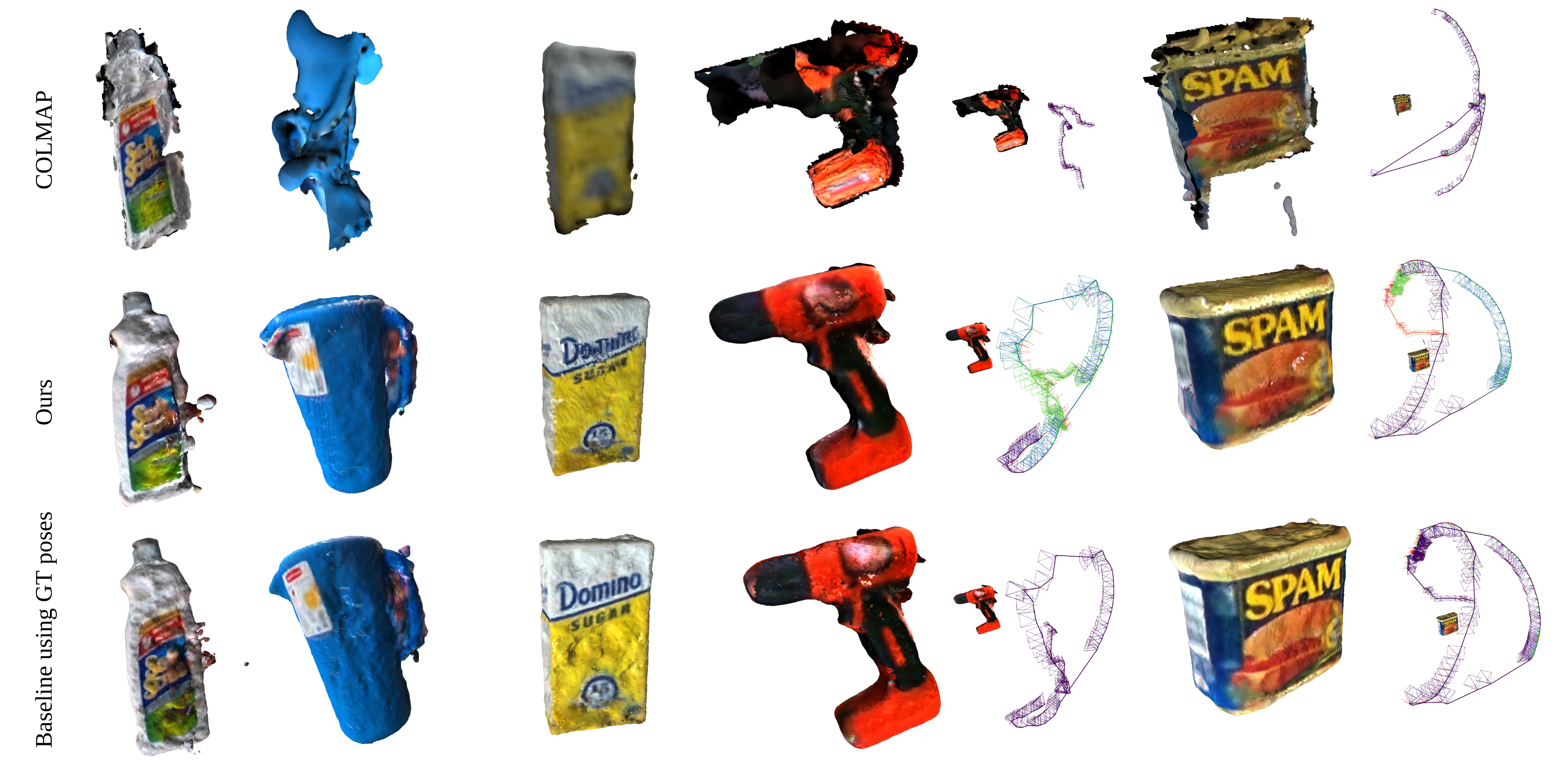}
\caption{\textbf{Reconstructed models and pose trajectories on HO-3D} for COLMAP~\cite{colmap}~(top row), our method~(middle row), and the UNISURF~\cite{unisurf} baseline that uses ground-truth poses~(bottom row). We restrict the keypoint matches in COLMAP to only object pixels using the segmentation masks obtained from a pre-trained network (Sec.~\ref{sec:reconstruction_setup}). COLMAP recovers only incomplete pose trajectories in absence of texture, which leads to incomplete or failed reconstructions. Our method relies on both geometric and texture features and produces reliable pose estimates, which results in similar reconstructions as produced by the strong baseline relying on ground-truth poses.
}
\label{fig:all_models}
\end{center}
\vspace{-2.5ex}
\end{figure*}

\section{Experiments} 
\label{sec:experiments}

We evaluate our method quantitatively and qualitatively on the HO-3D dataset~\cite{hampali-cvpr20-honnotate}, which contains sequences of objects from the YCB dataset~\cite{ycb} being manipulated by one hand. We also show qualitative results on the RGB images from the RGBD-Obj dataset~\cite{wang_dataset} and on sequences that we captured for this project and that show two challenging texture-less YCB objects: the clamp and the cube. The latter two datasets feature two hands but do not provide object pose annotations.
We evaluate the accuracy of the reconstructed shape and color, and of the estimated poses. 

\vspace{-1ex}
\customparagraph{3D Reconstruction Metric.} As in \cite{rgbd_tim}, we first align the estimated object mesh with the ground-truth mesh by ICP and then calculate the RMSE Hausdorff distance from the estimated mesh to the ground truth mesh. As our meshes are only estimated up to a scaling factor, we allow the meshes to scale during the ICP alignment.

\vspace{-1ex}
\customparagraph{Object Texture Metrics.} The recovered object texture is compared with the ground truth using the PSNR, LPIPS and SSIM metrics. Specifically, we render the recovered object appearance from the ground truth poses for images that were not used in the optimization and compare the renderings with the images. Since the pose has to be accurate to obtain reliable metrics, we first perform photometric optimization on the trained model to obtain accurate poses and then render the images as in BARF~\cite{barf}.

\vspace{-1ex}
\customparagraph{Pose Trajectory Metric.} As the 3D model and poses are recovered up to a 3D similarity transformation, we first align the estimated poses with the ground truth by estimating the transformation between the two.
We then calculate the absolute trajectory error (ATE)~\cite{david,imap,scnerf} and plot the percentage of frames for which the error is less than a threshold. We use the area under curve of the ATE plot as the metric.

\subsection{Evaluation on HO-3D}
HO-3D contains 27 multi-view sequences~(68 single-view sequences) of hand-object interactions with 10 YCB objects~\cite{ycb} annotated with 3D poses. We consider the same multi-view sequences as in \cite{rgbd_tim} for the 3D reconstruction. As the ground-truth 3D poses are provided in this dataset, we also evaluate the accuracy of our estimated poses along with the reconstruction and texture accuracy. 

\customparagraph{Baselines.} We compare with COLMAP~\cite{schonberger2016pixelwise}, the single-image object reconstruction method by Ye \etal \cite{cmu}, the RGB-D reconstruction method by Patten \etal \cite{rgbd_tim}, and UNISURF~\cite{unisurf}. The last two methods rely on the ground-truth camera poses, whereas the other methods (including ours) do not.
In the case of \cite{rgbd_tim}, we compare only the 3D reconstruction accuracy with this method as the pose and object texture evaluations are not reported. We obtain results from COLMAP using the sequential keypoint matching technique, and set the mesh trim parameter to 10. The largest connected component in the reconstructed mesh is then selected as the final result. We observed that COLMAP fails to obtain complete reconstruction for most objects due to insufficient keypoint matches and results in multiple non-overlapping partial reconstructions, which cannot be combined. The method by Ye \etal \cite{cmu} uses a single input image but is pre-trained on sequences of the same object. Note that our method is not pre-trained and thus the reconstructed object is completely unknown to the method.

\customparagraph{Results.} Table~\ref{tab:ho3d_model} compares the one-way RMSE Hausdorff distance of our method with COLMAP, the single-frame method of \cite{cmu}, and the RGBD method of \cite{rgbd_tim}. We calculate the average metric over the sequence for \cite{cmu}. Our method consistently achieves higher performance than COLMAP on all objects and higher performance over \cite{cmu} on average. Our method is competitive with the RGBD-based method and the strong baseline for most objects. COLMAP fails to obtain keypoint matches on banana and scissors. The lower accuracy of our method on pitcher and banana is due to the lack of both geometric and texture features.
COLMAP achieves accurate pose trajectories only for the cracker box and sugar box as they contain rich image features on all the surfaces. The lower accuracy of COLMAP on other objects is due to poor texture (Figure~\ref{fig:all_models}).

Table~\ref{tab:ate_area} shows the area under curve (AUC) of the absolute trajectory error (ATE) plot with the maximum ATE threshold of 10\,cm. Our method outperforms COLMAP, which cannot recover the complete trajectory for many objects. Both our method and COLMAP fail to obtain meaningful reconstruction for scissors due to its thin structure.

In Table~\ref{tab:psnr}, we provide the PSNR, SSIM and LPIPS metrics for our proposed method and the strong baseline that uses the ground-truth poses. Our method achieves similar accuracy as the baseline method on all objects, despite the fact that our method estimates the poses instead of using the ground-truth ones.
Results of our method, the UNISURF baseline and COLMAP are shown in Figure~\ref{fig:all_models}.

\begin{table}[]
    \footnotesize
    \begin{center}
    \begin{tabularx}{\columnwidth}{l Y Y Y | Y Y}
    \toprule
     Object  & Ye~\emph{et~al.} \cite{cmu} & COLMAP \cite{colmap} & Ours & UNISU. \cite{unisurf} & RGBD \cite{rgbd_tim}\\
    \midrule
    3: cracker box      &10.21& 4.08  & \textbf{2.91}  & 3.40 & 3.54\\
    4: sugar box        &6.19 & 6.66  & \textbf{3.01}  & 3.49 & 3.34\\
    6: mustard   &2.61 & \textbf{4.43}  & 4.44  & 4.34 & 3.28\\
    10: potted meat      &3.43 & 10.21 & \textbf{1.95}  & 1.54 & 3.26\\
    21: bleach  &\textbf{4.18} & 14.11 & 5.63  & 3.41 & 2.43\\
    35: power drill      &15.15& 11.06 & \textbf{5.48}  & 5.33 & 3.77\\
    19: pitcher base     &\textbf{8.87} & 43.38 & 9.21  & 4.63 & 4.73\\
    11: banana            &\textbf{3.47} & -     & 4.60  & 3.98 & 2.44\\
    \midrule
    Average                   &6.76 & 13.41 & \textbf{4.65} & 3.76 & 3.34\\
    \bottomrule
    \end{tabularx}
    \vspace{-1ex}
    \caption{\textbf{RMSE Hausdorff distance (mm) from the estimated to the ground-truth 3D model.} UNISURF~\cite{unisurf} and RGBD~\cite{rgbd_tim} are strong baselines as they use ground-truth poses, and the latter also depth images. Our object reconstructions are close to the baselines, even though we do not use the ground-truth poses, and systematically better than COLMAP, which slipped on the banana.}
    \label{tab:ho3d_model}
    \end{center}
    \vspace{-2.0ex}
\end{table}

\begin{table}
    \footnotesize
    \begin{center}
    \begin{tabularx}{\columnwidth}{l Y Y Y Y Y Y Y Y Y Y}
    \toprule
    Object & 3 & 4 & 6 & 10 & 21 & 35 & 19 & 25 & 11 & Avg \\
    
    \midrule
    COLMAP & 7.4 & \textbf{7.4} & 3.5 & 0.1 & 1.5 & 2.8 & 4.1 & \textbf{2.4} & 0.0 & 2.9\\
    Ours & \textbf{7.6}& 6.8 & \textbf{5.2} & \textbf{6.8} & \textbf{4.7} & \textbf{6.4} & \textbf{4.6} & 2.2 & \textbf{0.6} & \textbf{4.5}\\
    \bottomrule
    \end{tabularx}
    \vspace{-1ex}
    \caption{\textbf{Area under the curve of the absolute trajectory error.} COLMAP succeeds on textured objects like the first two but struggles to recover the complete trajectory for less textured objects.
    }
    \label{tab:ate_area}
    \end{center}
    \vspace{-2.0ex}
\end{table}
    
\begin{table}
    \footnotesize
    \begin{center}
    \begin{tabularx}{\columnwidth}{l Y Y}
    \toprule
    \multirow{2}{*}{\parbox{0.75cm}{\centering Object}} & \multicolumn{2}{c }{\centering PSNR$\uparrow$ ~/~SSIM$\uparrow$~/~LPIPS$\downarrow$}\\
    & Ours & UNISURF~\cite{unisurf}\\
    \midrule
    3: cracker box     & 29.77~/~0.73~/~0.31  & 29.79~/~0.74~/~0.33\\
    4: sugar box       & 30.77~/~0.82~/~0.31  & 30.73~/~0.76~/~0.33\\
    6: mustard  & 30.73~/~0.74~/~0.39  & 30.72~/~0.74~/~0.37\\
    10: potted meat     & 31.07~/~0.77~/~0.35  & 31.28~/~0.78~/~0.35\\
    21: bleach & 30.82~/~0.74~/~0.36  & 29.87~/~0.67~/~0.42\\
    35: power drill     & 31.82~/~0.78~/~0.26  & 31.81~/~0.76~/~0.28\\
    19: pitcher base    & 32.13~/~0.83~/~0.26  & 32.28~/~0.83~/~0.25\\
    25: mug              & 31.18~/~0.74~/~0.39  & 31.69~/~0.76~/~0.37\\
    \midrule
    Average                  & 31.01~/~0.77~/~0.32  & 31.02~/~0.75~/~0.34\\
    \bottomrule
    \end{tabularx}
    \vspace{-1ex}
    \caption{\textbf{Evaluation of the estimated object texture.} The proposed method achieves comparable quality of the recovered object texture as UNISURF which uses ground-truth poses.
    }
    \vspace{-2.0ex}
    \label{tab:psnr}
    \end{center}
\end{table}

\begin{table}
    \footnotesize
    \begin{center}
    \begin{tabularx}{\columnwidth}{l Y Y Y Y}
    \toprule
    Seq. Name & w/o $\loss_\dep$ & w/o $\loss_\opf$ & w/o $\loss_\reg$ & All Terms\\
    \midrule
    MDF14 & 4.9 & 4.8 & 1.1 & \textbf{5.8} \\
    SM2 & 1.0 & 2.9 & 0.5 & \textbf{8.1} \\
    \bottomrule
    \end{tabularx}
    \vspace{-1ex}
    \caption{\textbf{Ablation study with AUC of the ATE metric.} All loss terms are required for obtaining accurate trajectories.}
    \label{tab:abl_quant}
    \end{center}
    \vspace{-3.5ex}
\end{table}

\subsection{Evaluation on RGBD-Obj and New Sequences}

\begin{figure}
    \begin{center}
    \begin{minipage}{0.93\linewidth}
    \centering
    \subcaptionbox{Results on RGBD-Obj~\cite{wang_dataset}\label{fig:wang_result}}
    {\includegraphics[trim=220 325 380 77, clip,width=1\linewidth]{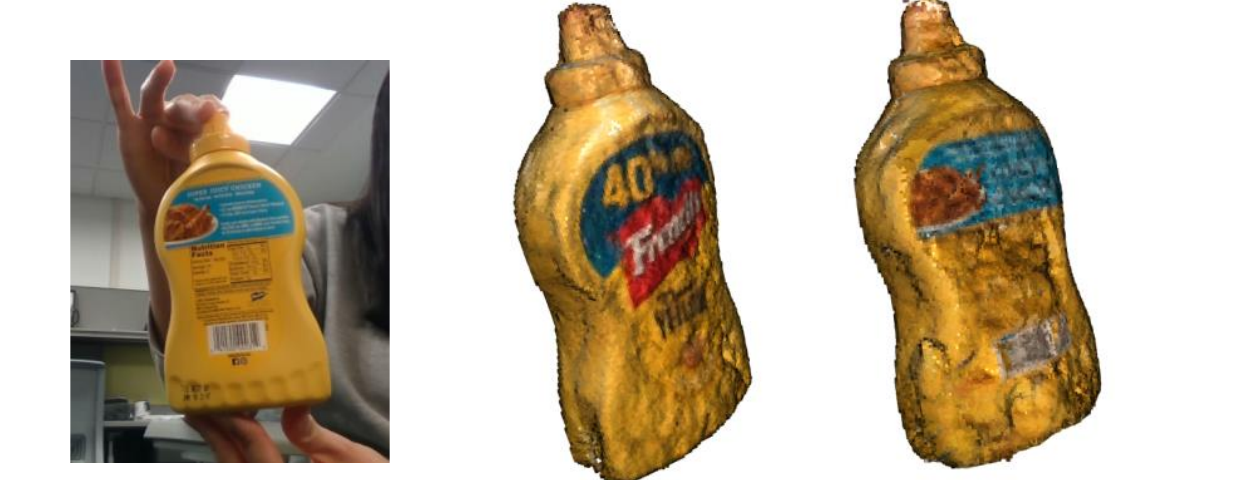}}
    \end{minipage}\vspace{3.0ex}
    \begin{minipage}{0.98\linewidth}
    \centering
    \subcaptionbox{Results on our new sequences\label{fig:our_dataset_results}}
    {\includegraphics[trim=220 70 380 225, clip,width=1\linewidth]{figures/main_results/wang_ours_results.pdf}}
    \end{minipage}
    \vspace{0.5ex}
    \caption{\textbf{Results on RGBD-Obj~\cite{wang_dataset} and two new sequences.} The left column shows a sample image from the input sequence showing an unknown object manipulated by hands. The right column shows two views at the reconstructed color 3D model.}
    \label{fig:new_dataset_results}
    \end{center}
    \vspace{-2.5ex}
\end{figure}

Qualitative results from an RGBD-Obj~\cite{wang_dataset} sequence showing the mustard bottle and from two new sequences with the extra large clamp and Rubik's cube from YCB, which we captured for this project, are shown in Figure~\ref{fig:new_dataset_results}. Our method is able to produce 3D models also for the latter two objects, which are poorly textured and classical feature-based methods such as~\cite{schonberger2016pixelwise} fail to reconstruct them.


\subsection{Ablation Study}

The benefit of individual loss terms proposed in Sec.~\ref{sec:opt_obj} is demonstrated qualitatively in Figures~\ref{fig:optical_flow}--\ref{fig:depth_term} and quantitatively in Table~\ref{tab:abl_quant}, where AUC of the ATE plot is shown for the largest segment for 2 sequences from the HO-3D dataset. We do not consider complete reconstruction in Table~\ref{tab:abl_quant} as tracking fails completely for some segments without some of the loss terms. The optical flow loss enforces provides additional constraints on the predicted camera poses, the shape regularization loss stabilizes the optimization especially in its early stage, and the synthetic depth loss preserves previously reconstructed surface parts. Without the synthetic depth loss, the object can be significantly deformed especially when the camera performs larger motions in newly considered frames.

Figure~\ref{fig:seg_ablation} shows the importance of splitting the input sequence into segments. Figure~\ref{fig:rand_init} demonstrates the benefit of initializing the optimization of Eq.~\eqref{eq:complete_loss} with poses estimated from segments over initializing with random or zero poses.

\section{Conclusion}

We introduced a method that is able to reconstruct an unknown object manipulated by hands from color images.
The main challenge resides in preventing drift during the simultaneous tracking and reconstruction. We believe our strategy of splitting the sequence based on the apparent area of the object and our regularization terms to be general and useful ideas, and hope they will inspire other researchers.



{\small
\bibliographystyle{ieee_fullname}
\bibliography{vincents_refs,egbib}
}

\end{document}



\title{In-Hand 3D Object Scanning from an RGB Sequence\\~\\Supplementary Material\\}
\newcommand{\namesep}{\hspace{0.8em}}
\author{Shreyas Hampali\textsuperscript{1,3}\thanks{Work done as part of Shreyas's PhD thesis at TU Graz, Austria.}~~\namesep
Tomas Hodan\textsuperscript{1}\namesep
Luan Tran\textsuperscript{1}\\
Lingni Ma\textsuperscript{1}\namesep
Cem Keskin\textsuperscript{1}\namesep
Vincent Lepetit\textsuperscript{2,3} \and
\textsuperscript{1}{\normalsize Reality Labs at Meta}\\ 
\textsuperscript{2}{\normalsize 
LIGM, Ecole des Ponts, Univ Gustave Eiffel, CNRS, Marne-la-Vall\'ee, France
}\\
\textsuperscript{3}{\normalsize Institute for Computer Graphics and Vision, Graz University of Technology, Graz, Austria}}

\maketitle

\renewcommand{\reg}{\text{r}}

We provide more details and additional results of our method in this supplementary material and in the attached supplementary video. We discuss results on sequences captured with the Aria glasses, additional results on the HO-3D dataset, and describe the limitations of our approach.

\section{In-hand Object Scanning with Aria~\cite{aria_pilot_dataset}}
\label{sec:aria}

\begin{figure}[b]
    \centering
    \includegraphics[trim=100 180 205 240,clip,width=1\linewidth]{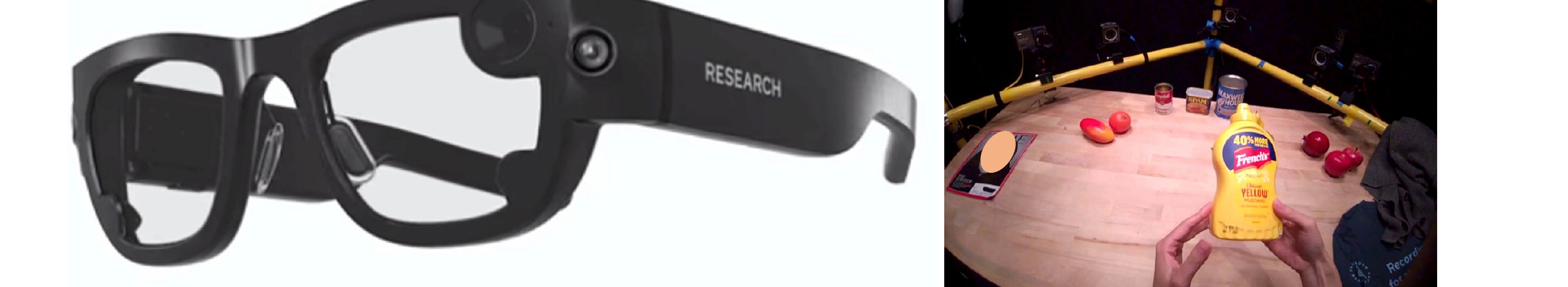}
    \caption{\textbf{Aria glasses.} Aria glasses~\cite{aria_pilot_dataset} provide egocentric views of the environment using cameras mounted on the glasses.}
    \label{fig:aria_ex}
\end{figure}

The recently introduced Aria AR glasses~\cite{aria_pilot_dataset} provide a first-person capture of the environment using cameras mounted on the glasses. A head-mounted camera provides an intuitive and simple way for scanning an object using both hands.
Here we show that our method can be applied for reconstruction of unknown objects from sequences captured using the Aria glasses. Figure~\ref{fig:aria_ex} shows the Aria glasses and an egocentric view of two hands manipulating an object from the YCB dataset.

\begin{figure*}
    \centering
    \includegraphics[trim=200 180 205 140,clip,width=1\linewidth]{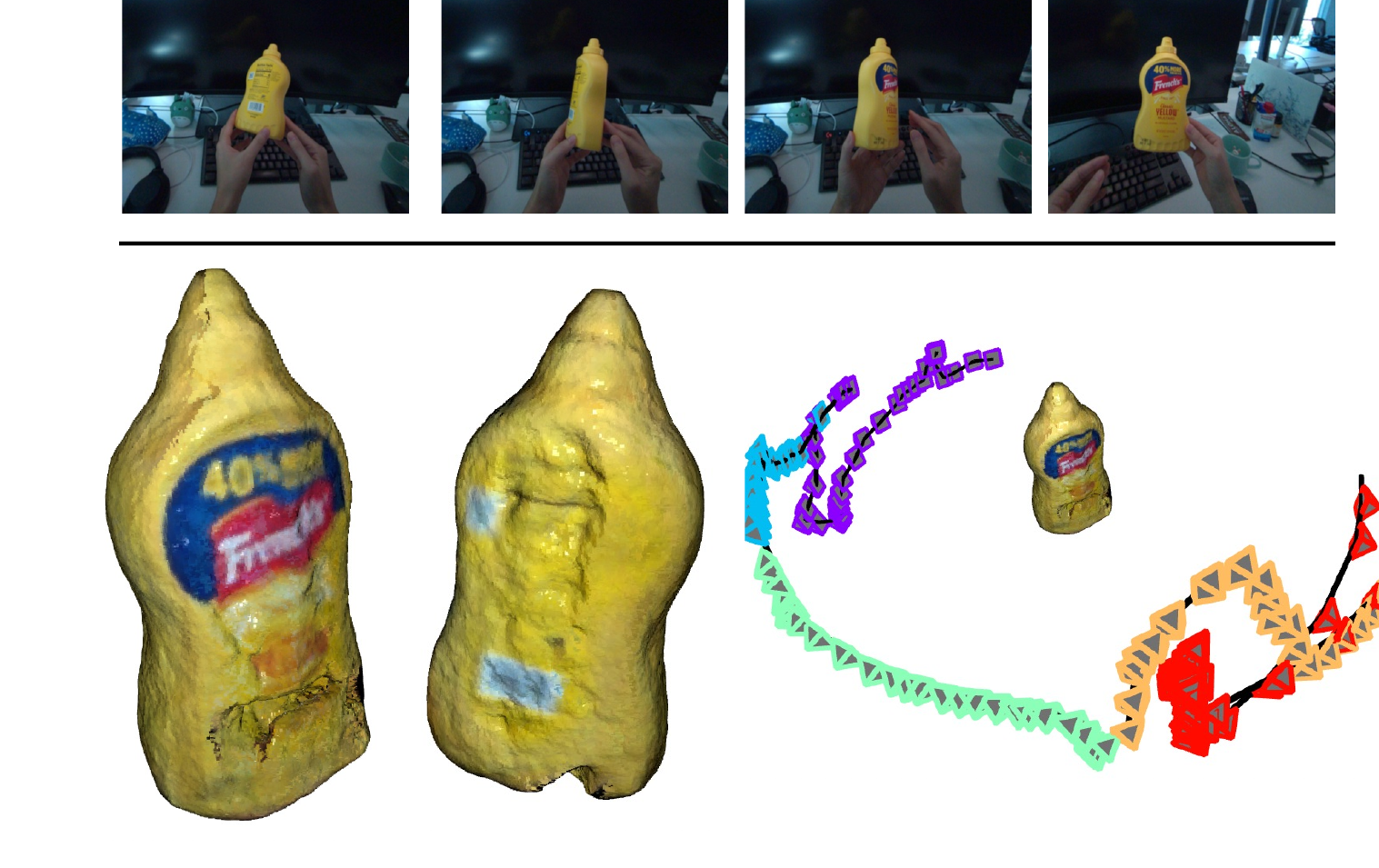}
    \caption{\textbf{In-hand object scanning with Aria glasses.} Our method can be applied to reconstruct (bottom left) objects and estimate its poses (bottom right) from RGB sequences captured using Aria glasses (top). The different segments created by our approach is color coded in the pose trajectory.}
    \label{fig:aria_res}
\end{figure*}

We first linearize the fish-eye images from the Aria sequence and use Detic~\cite{zhou2022detecting} to obtain the hand and unknown object masks in the images. The reconstruction result on the mustard bottle sequence captured using the Aria glasses shown in Figure~\ref{fig:aria_res} demonstrates that our proposed method can be applied to this in-hand scanning scenario as well.

\section{More Qualitative Results}
\begin{figure*}
    \centering
    \includegraphics[page=1,trim=10 80 10 60,clip,width=0.9\linewidth]{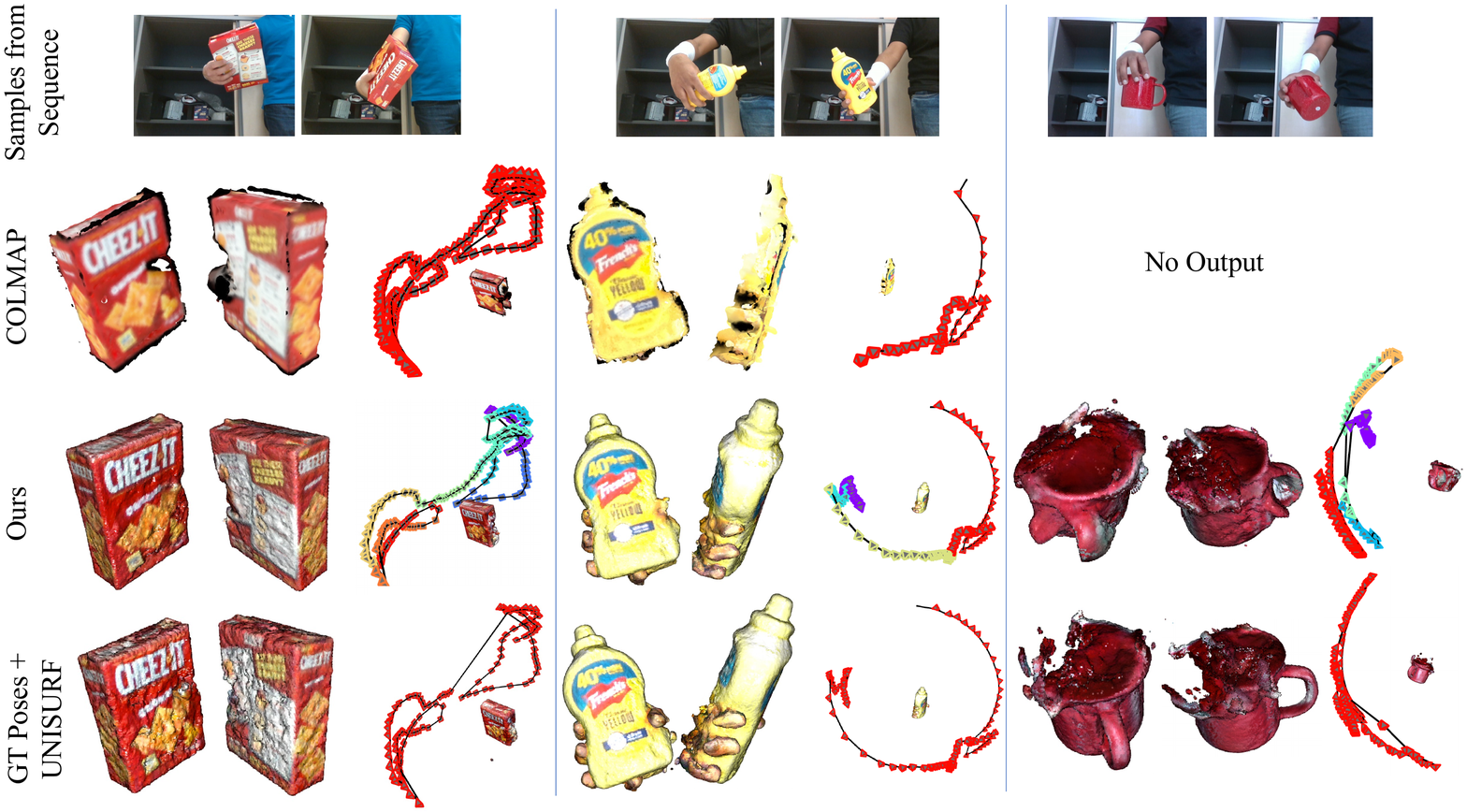}
    \caption{\textbf{More qualitative results (reconstructed models and pose trajectories) on HO3D dataset.} Left column: Our method and COLMAP obtain high quality reconstruction on textured objects. Middle column: Our method manages to return a complete reconstruction on this partially textured object, while COLMAP fails to reconstruct the back. The fingers reconstructed as part of the mustard bottle are due to inaccurate hand masks.  Third column: We achieve reasonable results on this very challenging texture-less object, on which COLMAP fails completely. We could not reconstruct the parts of the object that are always occluded by the hand. The reconstruction quality of our method is similar to the quality obtained when using the ground truth poses.}
    \label{fig:suppl_results}
\end{figure*}

We show more qualitative results from the HO-3D sequences in Figure~\ref{fig:suppl_results}. Corresponding quantitative results are provided in Tables~1-3 of the main paper. Our method can reconstruct partially-textured and texture-less objects such as the mustard bottle and mug in Figure~\ref{fig:suppl_results}. Fingers grasping the object are reconstructed on the mustard bottle sequence (second column of Figure~\ref{fig:suppl_results}) due to inaccurate hand masks and static grasp pose of the hand throughout the sequence. Parts of the object that are always occluded by the hand as in the mug sequence (third column of Figure~\ref{fig:suppl_results}) are also inaccurately reconstructed as we do not assume any prior on the object shape.

\section{Shape Regularization Loss}
Minimizing the shape regularization loss (Eq.~9 of the main paper) at $t=0$ encourages the reconstructed object surface to be a plane parallel to the image plane. This can be seen by considering an orthographic projection of the rays and noting that for each object ray $\ray$ that passes through an object pixel in the camera at $t=0$,

\begin{align}
    \min \sum_k o_\theta(\bx_k) \exp{\big(\alpha \cdot \|\bx_k\|_2\big)} &= \min_k \exp{(\alpha \cdot \|\bx_k\|_2)} \nonumber \\
    &= \exp{\alpha \cdot \big(\min_k \|\bx_k\|_2\big)} \nonumber
\end{align}
where the first equality is due to $o_\theta(\bx_k) = \{0,1\} \forall k$. As we assume orthographic projection, the point on the object ray with the minimum euclidian distance to the origin lies in a plane that is parallel to the image plane and passing through the origin. Thus, the for all the object rays the reconstructed surface is on this plane.


\section{More Implementation Details}

\begin{figure*}[t]
    \centering
    \begin{minipage}{0.7\linewidth}
        \subcaptionbox{Bleach bottle}
        {\includegraphics[page=1,trim=170 100 200 120,clip,width=1\linewidth]{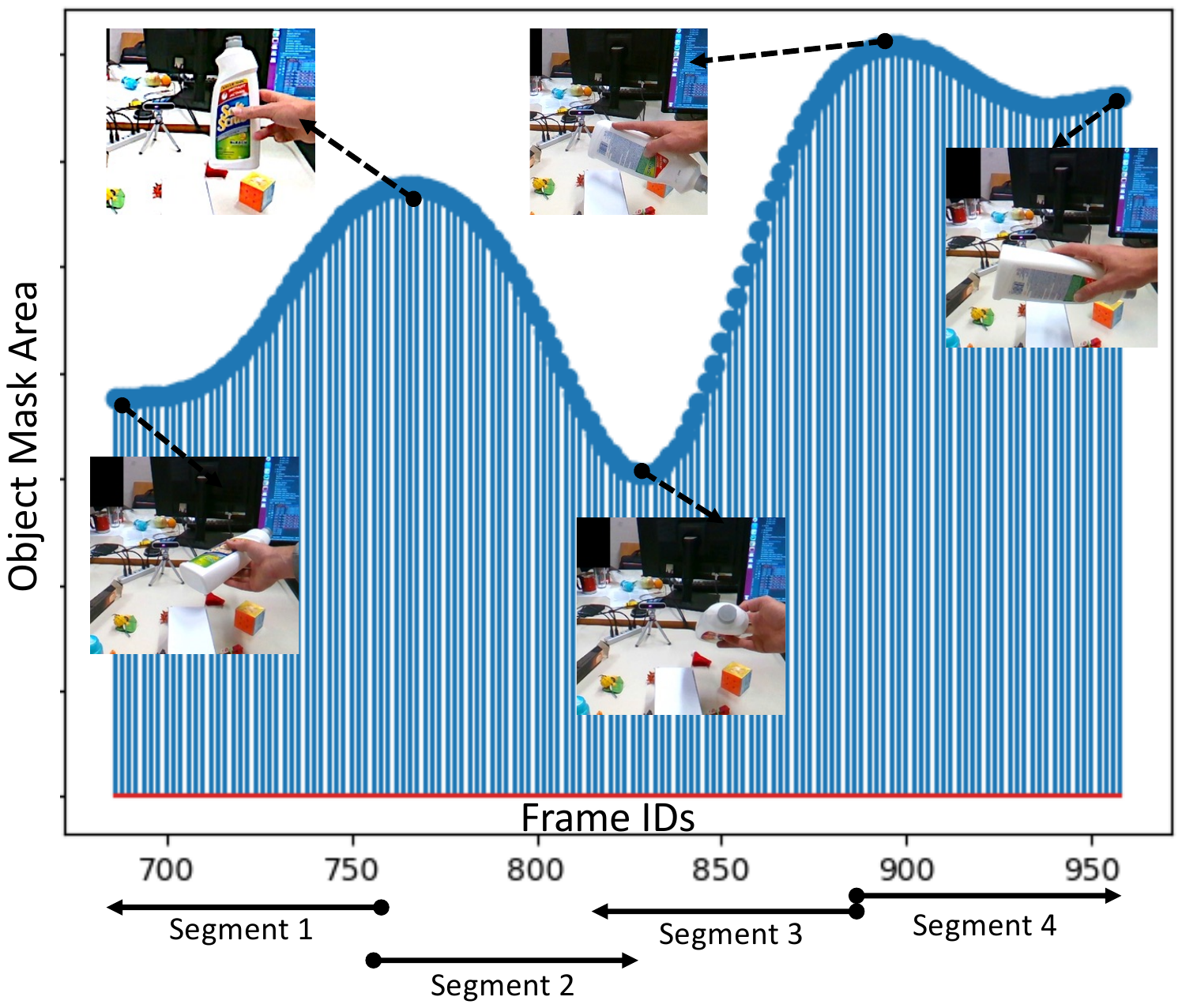}}
    \end{minipage}
    \begin{minipage}{0.7\linewidth}
        \subcaptionbox{Pitcher base}
        {\includegraphics[page=2,trim=175 110 235 140,clip,width=1\linewidth]{figures/supplementary/area_curves.pdf}}
    \end{minipage}~
    \caption{\textbf{Object area curves and segment boundaries.} We show the segment boundaries for two objects (bleach bottle and pitcher base) which are calculated from the object area curves. In each segment, the incremental object reconstruction and pose tracking starts at the local maximum of the object area and ends at the local minimum.}
    \label{fig:obj_area}
\end{figure*}

As discussed in Section~3.4.1 of the main paper, we divide the input RGB sequence into multiple overlapping segments, and incrementally reconstruct the object shape and estimate its pose in each segment. As reconstructing the object from every frame of the entire RGB sequence is not feasible, we first subsample the input RGB sequence and manually select the frame interval from the entire sequence on which we run our method. The interval is selected such that all parts of the object are visible during scanning. In Table~\ref{tab:ho3d_seq_info}, we provide the names of the sequences from the HO-3D dataset which are used for reconstruction, the chosen frame intervals, and the number of segments the RGB sequence is divided into by our method. Additionally, in Figure~\ref{fig:obj_area}, we show the frame interval on which the reconstruction is performed for two objects in the HO-3D dataset along with the segment boundaries

\begin{figure}
    \centering
    \begin{minipage}{0.33\linewidth}
        \subcaptionbox{Image}
        {\includegraphics[page=1,trim=100 200 620 120,clip,width=1\linewidth]{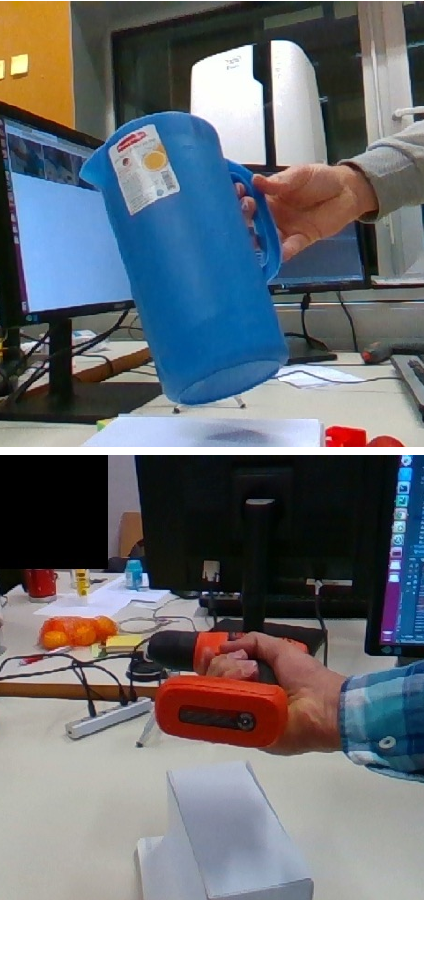}}
    \end{minipage}~
    \begin{minipage}{0.33\linewidth}
        \subcaptionbox{Object mask}
        {\includegraphics[page=1,trim=275 200 445 120,clip,width=1\linewidth]{figures/supplementary/masks.pdf}}
    \end{minipage}~
    \begin{minipage}{0.33\linewidth}
        \subcaptionbox{Hand mask}
        {\includegraphics[page=1,trim=455 200 265 120,clip,width=1\linewidth]{figures/supplementary/masks.pdf}}
    \end{minipage}
    \caption{\textbf{Hand and object segmentation masks.} We obtain foreground masks from \cite{boerdijk2020learning} and hand masks from \cite{wu2021seqformer}.}
    \label{fig:hand_obj_masks}
\end{figure}

\begin{figure}
    \centering
    \includegraphics[page=1,trim=160 120 120 100,clip,width=1\linewidth]{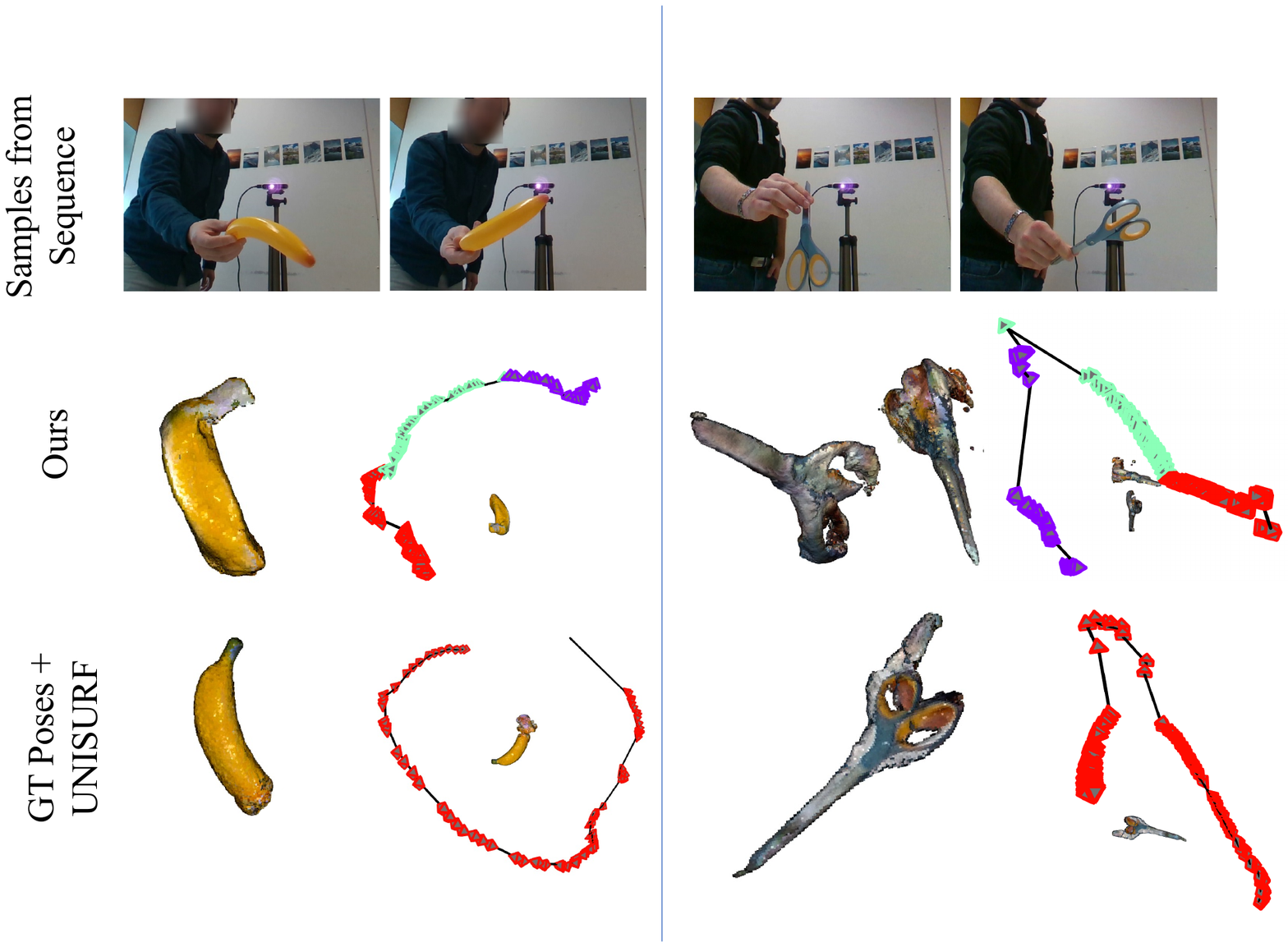}
    \caption{\textbf{Failure scenarios.} Our method fails to obtain poses and reconstruct texture-less symmetrical (left) and thin objects (right).}
    \label{fig:limitations}
    \vspace{-2ex}
\end{figure}

\begin{table}[t!]
    \footnotesize
    \begin{center}
    \begin{tabularx}{\columnwidth}{l Y Y Y Y}
    \toprule
    Object  & Sequence ID & Start Frame ID & End Frame ID & No. of Segments\\
    \midrule
    3: cracker box       &MC1    & 210 & 650 & 8\\
    4: sugar box         &ShSu12 & 276 & 1552& 8\\
    6: mustard           &SM2    & 300 & 576 & 4\\
    10: potted meat      &GPMF12 & 70  & 334 & 4\\
    21: bleach           &ABF14  & 686 & 960 & 4\\
    35: power drill      &MDF14  & 270 & 772 & 3\\
    19: pitcher base     &AP13   & 60  & 370 & 4\\
    11: banana           &BB12   & 1100&1260 & 4\\
    25: mug              &SMu1   & 702 &1236 & 5\\
    \bottomrule
    \end{tabularx}
    \vspace{-1ex}
    \caption{\textbf{Sequence IDs and frame intervals chosen for reconstruction from the HO-3D dataset, and number of segments created by our approach.}}
    \label{tab:ho3d_seq_info}
    \end{center}
    \vspace{-3.5ex}
\end{table}

\section{Hand and Object Masks}
We show some object and hand masks used by our method in Figure~\ref{fig:hand_obj_masks}. We rely on the pre-trained network from \cite{boerdijk2020learning} which segments dynamic foreground from static background and segments hand and object as one class. We then obtain hand-only masks from \cite{wu2021seqformer} and combine with foreground mask from \cite{boerdijk2020learning} to obtain hand and object masks. 

For the Aria sequences, as discussed in Section~\ref{sec:aria} which do not have static background, we use Detic~\cite{zhou2022detecting} to obtain hand and object masks.

\section{Limitations}

Our method relies on the geometric or texture features on the object to incrementally reconstruct and estimate its pose within a segment.The proposed approach results in inaccurate pose estimates for texture-less and nearly symmetrical objects such as banana leading to erroneous reconstruction as shown in Figure~\ref{fig:limitations}. Our method also fails to estimate poses of thin objects such as scissors leading to inaccurate reconstructions as also shown in Figure~\ref{fig:limitations}. We believe hand pose information can provide additional cues to estimate the object poses during more challenging scenarios and is a potential future direction for our approach.

{\small
\bibliographystyle{ieee_fullname}
\bibliography{egbib}
}